\documentclass[letterpaper, 10 pt, conference]{ieeeconf}  

\IEEEoverridecommandlockouts                              

\usepackage{graphicx}
\usepackage{amsmath} 
\newtheorem{problem}{Problem}
\newtheorem{theorem}{Theorem}
\newtheorem{lemma}{Lemma}

\usepackage{amssymb}  
\usepackage{algorithm}
\usepackage{algpseudocode}
\usepackage{xcolor}
\algnewcommand\algorithmicinput{\textbf{INPUT:}}
\algnewcommand\INPUT{\item[\algorithmicinput]}
\algnewcommand\algorithmicoutput{\textbf{OUTPUT:}}
\algnewcommand\OUTPUT{\item[\algorithmicoutput]}
\usepackage{subfiles}

\usepackage[font=footnotesize,belowskip=-10pt,aboveskip=3pt]{caption}

\renewcommand{\baselinestretch}{0.92}

\title{\LARGE \bf
Risk-Sensitive Sequential Action Control with Multi-Modal Human Trajectory Forecasting for Safe Crowd-Robot Interaction 
}

\author{Haruki Nishimura$^{1}$, Boris Ivanovic$^{1}$, Adrien Gaidon$^{2}$, Marco Pavone$^{1}$, and Mac Schwager$^{1}$
\thanks{Toyota Research Institute ("TRI")  provided funds to assist the authors with their research but this article solely reflects the opinions and conclusions of its authors and not TRI or any other Toyota entity. This work was also supported in part by ONR grant number N00014-18-1-2830, a JASSO fellowship, and a Masason Foundation fellowship. The authors are grateful for this support.}
\thanks{$^{1}$Haruki Nishimura, Boris Ivanovic, Marco Pavone, and Mac Schwager are with the Department of Aeronautics and Astronautics, Stanford University, Stanford, CA 94305 USA
        {\tt\small  \{hnishimura, borisi, pavone, schwager\}@stanford.edu}}%
\thanks{$^{2}$Adrien Gaidon is with Toyota Research Institute, Los Altos, CA 94022 USA
        {\tt\small adrien.gaidon@tri.global}}%
}

\begin{document}

\maketitle
\thispagestyle{empty}
\pagestyle{empty}

\begin{abstract}

This paper presents a novel online framework for safe crowd-robot interaction based on risk-sensitive stochastic optimal control, wherein the risk is modeled by the entropic risk measure. The sampling-based model predictive control relies on mode insertion gradient optimization for this risk measure as well as Trajectron++, a state-of-the-art generative model that produces multimodal probabilistic trajectory forecasts for multiple interacting agents. Our modular approach decouples the crowd-robot interaction into learning-based prediction and model-based control, which is advantageous compared to end-to-end policy learning methods in that it allows the robot's desired behavior to be specified at run time. In particular, we show that the robot exhibits diverse interaction behavior by varying the risk sensitivity parameter. A simulation study and a real-world experiment show that the proposed online framework can accomplish safe and efficient navigation while avoiding collisions with more than 50 humans in the scene.
 \end{abstract}

\section{INTRODUCTION}
\label{sec: intro}

As autonomous robots expand their workspace to cage-free, social environments, they must be designed as safety-critical systems; failures in avoiding collisions with humans sharing the workspace result in catastrophic accidents. Safe and efficient robot navigation alongside many humans is still a challenging problem in robotics, especially due to the potentially unpredictable and uncooperative nature of human motion. 
We propose an effective solution to this problem via risk-sensitive stochastic optimal control, wherein desired collision avoidance and goal reaching motion is achieved by a cost function, a risk-sensitivity parameter, and dynamic optimization. Specifically, we extend the Stochastic Sequential Action Control (SAC) algorithm \cite{nishimura2018sacbp} to a risk-sensitive setting through the use of exponential disutility \cite{whittle2002risk}, the objective often referred to as the entropic risk measure \cite{majumdar2020should}. The proposed sampling-based algorithm, which we name Risk-Sensitive Sequential Action Control (RSSAC), is a stochastic nonlinear model predictive control (NMPC) algorithm that optimally improves upon a given nominal control with a series of control perturbations. Unlike many other control-theoretic stochastic NMPCs \cite{todorov2005generalized, gomez2016real, williams2018information}, RSSAC is not limited to a particular class of distributions such as Gaussian, nor does it need to know the analytical form; it only requires a black-box probabilistic generator. 
Leveraging this property and recent advances in machine learning for modeling multi-agent behavior, we combine RSSAC with Trajectron++ \cite{salzmannivanovic2020trajectron++}, a state-of-the-art generative model for predicting the many possible future trajectories of multiple interacting agents accurately and efficiently. The overall framework constitutes a prediction-control pipeline for safe robot navigation, which is shown to be capable of interacting with more than 50 pedestrians simultaneously while avoiding collisions and steering the robot towards its goal. 

\begin{figure}
    \centering
    \scalebox{0.9}[0.9]{\includegraphics[clip,width=1.0\columnwidth]{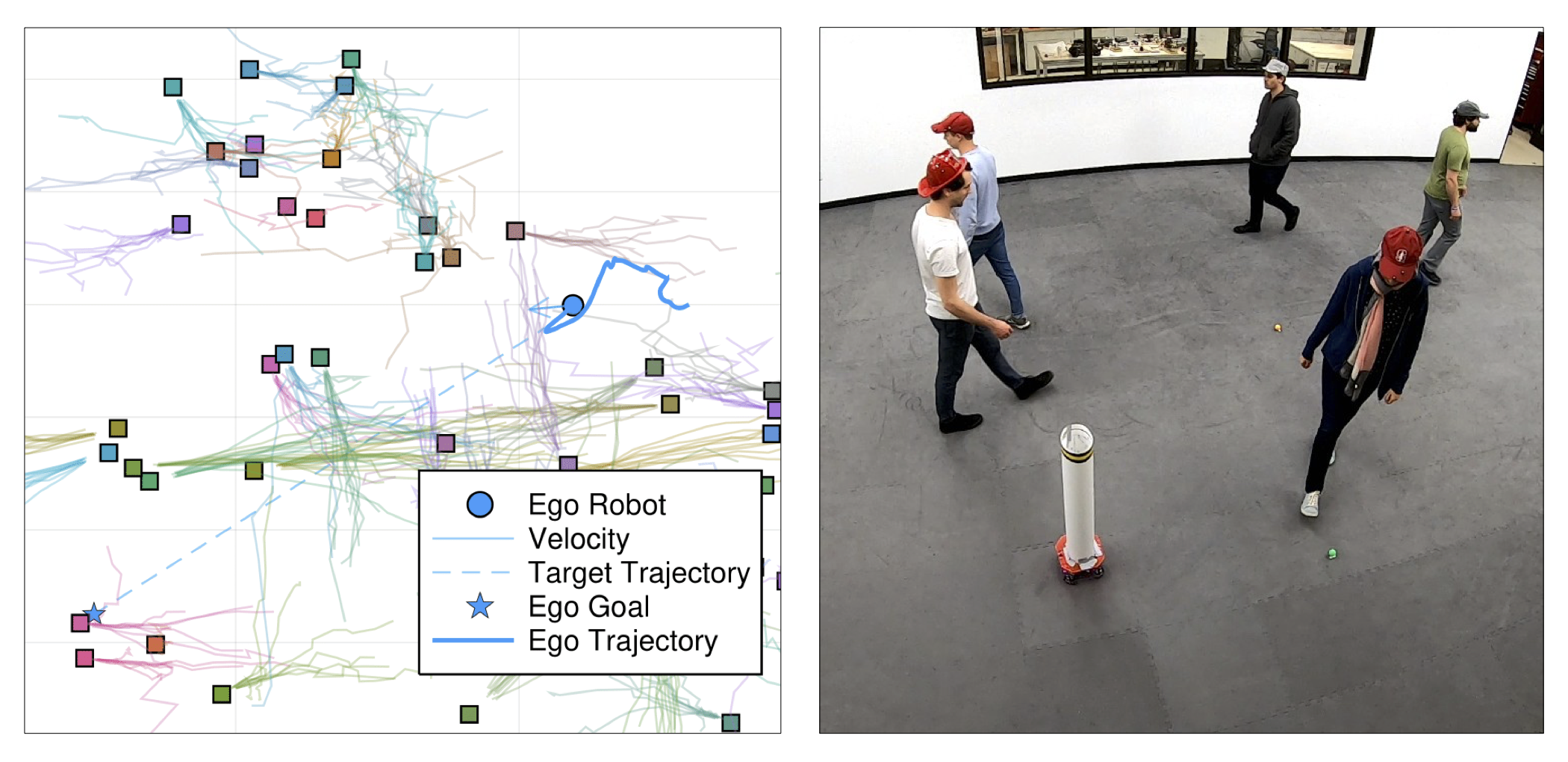}}
    \caption{The proposed RSSAC-Trajectron++ framework is effective for safe robot navigation in a social environment densely populated with humans. (Left) A simulation environment with real human trajectories from the UCY/UNIV scene \cite{lerner2007crowds}, overlaid with predictions sampled from Trajectron++. (Right) A risk-sensitive robot safely navigating itself alongside 5 humans.}
    \label{fig: demo}
\end{figure}

The contributions of this paper are two-fold. First, the extension of the Stochastic SAC to risk-sensitive optimal control is presented as a theoretical contribution. This is achieved by generalizing the mode insertion gradient \cite{wardi2012algorithm, ansari2016sequential} to the entropic risk measure. Second, the combined RSSAC-Trajectron++ framework is presented as a novel modular approach to safe crowd-robot interaction with dynamic collision avoidance. RSSAC takes advantage of the parallelizability of Monte Carlo sampling, leading to an efficient GPU implementation with real-time performance. We conduct simulation studies as well as a real-world experiment to demonstrate that the explicit probabilistic prediction of Trajectron++ and risk-sensitive optimization of RSSAC allow a robot to interact with many humans safely and efficiently. They also reveal that the risk sensitivity parameter adds an additional degree of freedom in determining the type of interaction behavior exhibited by the robot, e.g. yielding to or crossing in front of oncoming humans; this diverse behavior is not achieved by tuning cost function parameters.

\section{RELATED WORK}
\label{sec: related}

We first review relevant work in dynamic collision avoidance and safe robot navigation in Section \ref{sec: related_work_col_avoidance}. We then focus on Stochastic SAC (Section \ref{sec: related_work_sac}) and data-driven trajectory forecasting methods (Section \ref{sec: related_work_trajectron}), on which our RSSAC-Trajectron++ framework is built.

\subsection{Dynamic Collision Avoidance and Safe Robot Navigation}
\label{sec: related_work_col_avoidance}
There exists a vast body of literature in the field of dynamic collision avoidance and safe robot navigation. Coordinated collision avoidance is widely studied in the multi-robot systems literature, with methods including 
mixed integer programming \cite{mellinger2012mixed}, reciprocal velocity obstacles \cite{van2011reciprocal}, buffered Voronoi cells \cite{zhou2017fast, wang2018safe}, Nash-equilibria-based differential games \cite{mylvaganam2017differential}, 
and safety barrier certificates \cite{wang2017safety}. These methods do not explicitly model uncoordinated agents or motion uncertainty, which are both crucial aspects in human motion; later in Section \ref{sec: results} we observe collisions for \cite{wang2018safe} due to humans breaking the coordination assumption. A notable exception is \cite{luo2019multi}, where the authors propose probabilistic safety barrier certificates for bounded motion uncertainty. An alternative approach suited to bounded uncertainty is the Hamilton-Jacobi (HJ) reachability analysis \cite{tomlin1998conflict, chen2016multi}, although it suffers from the curse of dimensionality in solving the Hamilton-Jacobi-Isaacs equation.

Another common approach to modeling motion uncertainty is via the use of probability distributions. MDP and POMDP frameworks have been applied to robot navigation and collision avoidance \cite{temizer2010collision, cai2018hyp} and are best suited to discrete action spaces. Stochastic Optimal Control (SOC), especially sampling-based model predictive control (MPC), is also gaining increasing attention due to its ability to handle stochastic uncertainty and incorporate prediction of possible future outcomes. Recent methods applied to safe robot navigation and dynamic collision avoidance include exhaustive search with motion primitives \cite{schmerling2018multimodal}, path-integral control \cite{gomez2016real}, and information-theoretic control \cite{williams2018best}. While the latter two approaches can find approximately optimal solutions under certain assumptions, they require the form of distribution to be Gaussian. In addition, it is often not satisfactory to only optimize for the expected cost, especially for safety-critical systems. There exist many possible ways to endow robots with risk-awareness \cite{majumdar2020should}, such as chance-constrained optimization \cite{blackmore2011chance}, conditional value at risk (CVaR) \cite{samuelson2018safety}, and entropic risk measure \cite{medina2012risk}. All of these methods come with advantages and drawbacks, but in general there is a trade-off between the complexity of modeling assumptions and the statistical safety assurances. Our work employs the entropic risk measure which has been extensively studied in control theory \cite{whittle1981risk, glover1988relations}. 
Albeit not yielding statistical safety assurances, our algorithm 
can compute control inputs in real-time for highly dynamic environments with multimodal predictive distributions, which is quite challenging for chance constrained optimization or CVaR-based methods.

Lastly, we briefly highlight similarities and differences of our work from policy learning methods such as imitation learning \cite{pokle2019deep} or end-to-end deep reinforcement learning \cite{everett2018motion,chen2019crowd}. Similar to these methods, our framework applies machine learning to extract features from multi-agent interactions. However, we use these features to make explicit probabilistic predictions about other agents which are then incorporated into optimization by the model-based controller. Our modular approach has two advantages. First, explicit prediction brings about better interpretability and explainability of the resulting interaction. Second, our model-based controller allows the robot's behavior to be changed at run time by simply modifying the cost function design or the risk sensitivity parameter, while other methods would need re-training to do so as the policy itself is data-driven.

\subsection{Stochastic Sequential Action Control}
\label{sec: related_work_sac}
SAC in its original form is a deterministic model predictive control algorithm for nonlinear hybrid systems \cite{ansari2016sequential, tzorakoleftherakis2018iterative}. In contrast to local trajectory optimization, at each time the algorithm perturbs a given (open-loop or closed-loop) nominal control to optimize a local improvement of the total receding-horizon cost, a quantity known as the mode insertion gradient \cite{wardi2012algorithm}. SAC is shown to yield highly efficient control computation with high-frequency replanning, which often achieves better performance than local trajectory optimization methods. In prior work \cite{nishimura2018sacbp, nishimura2019sacbp}, we have generalized the mode insertion gradient to stochastic systems and solved challenging belief space planning problems, where our approach significantly outperforms other state-of-the-art methods under both sensing and motion uncertainty. The proposed RSSAC algorithm is an extension of this algorithm to risk-sensitive optimal control in which the cost functional is the entropic risk measure. This makes the algorithm more suitable for problems where safety is concerned, such as the crowd-robot interaction one that we address in this paper.

\subsection{Multi-Agent Trajectory Modeling from Data}
\label{sec: related_work_trajectron}
Since human behavior is rarely deterministic or unimodal, deep learning-based generative approaches have emerged as state-of-the-art trajectory forecasting methods, due to recent advancements in deep generative models \cite{SohnLeeEtAl2015,GoodfellowPouget-AbadieEtAl2014}. 
They mostly use a recurrent neural network architecture with a latent variable model, such as a Conditional Variational Autoencoder (CVAE) \cite{SohnLeeEtAl2015}, to explicitly encode multimodality (e.g.~\cite{ivanovic2019trajectron}),
or a Generative Adversarial Network (GAN) \cite{GoodfellowPouget-AbadieEtAl2014} to do so implicitly (e.g.~\cite{GuptaJohnsonEtAl2018}).
Trajectron++ \cite{salzmannivanovic2020trajectron++} belongs to the former, making use of a CVAE to explicitly model the multimodality of human behavior with a latent variable. Specifically, it uses a discrete Categorical latent variable which aids in performance as well as interpretability, since different behavior modes are explicitly represented and able to be visualized. Designed for downstream robotic motion planning and control tasks, Trajectron++ can also produce predictions that are conditioned on candidate future motion plans of the ego robot. We leverage this capability in a real-world experiment in Section \ref{sec: results} and show that the robot-future-conditional prediction significantly improves both safety and efficiency of the robot navigation.

\section{PROBLEM STATEMENT}
\label{sec: problem}

\subsection{Dynamics Model}
Following \cite{nishimura2018sacbp, nishimura2019sacbp}, this work assumes a discrete-time dynamical system model for other agents (i.e. humans) while employing a continuous-time one for the robot. This modeling assumption is practical, as 1) in general the robot receives information of its surrounding environment at a much lower frequency than actuation; and 2) with this representation one can naturally handle potentially agile robot dynamics without coarse time discretization. Let $x(t) \in \mathbb{R}^n$ denote the state of the robot at time $t$, whose dynamics are modeled by
\begin{align}
    \label{eq: robot_dynamics}
    \dot{x}(t) = f(x(t)) + H(x(t))u(t),
\end{align}
where $u(t) \in \mathcal{U} \subseteq \mathbb{R}^m$ is the control and $\mathcal{U}$ is a bounded convex set. We assume that the dynamics are deterministic and control-affine. 
As the robot navigates in the environment, it receives position information on humans in the scene and updates it at discrete times $\{t_k\}_{k \geq 0}$, with interval time $\Delta t_\text{o}$. Similarly to related works \cite{zhou2017fast, wang2018safe} we do not assume velocity information to be available to the robot. Let $p^{i}(t_k) \in \mathbb{R}^2$ denote the position of the human with label $i \in \{1, \dots, N\}$. 
From $t_{k-1}$ to $t_{k}$, the position $p^i(t_{k-1})$ changes to $p^i(t_{k})$ with difference $y^i_{k}$. From the robot's continuous-time perspective,
this is viewed as a periodic jump discontinuity given by
\begin{align}
    \label{eq: human_dynamics}
    \begin{cases}
        p^i(t_k) = p^i(t_k^-) + y^i_k &~\\
        p^i(t) = p^i(t_k) &\forall t \in [t_k, t_{k+1}),
    \end{cases}
\end{align}
where $t_k^-$ is infinitesimally smaller than $t_k$.
The dynamics of the joint state $s(t) = (x(t), p^{1}(t), \dots, p^{N}(t))$ are specified by \eqref{eq: robot_dynamics} and \eqref{eq: human_dynamics}, which together constitute a hybrid dynamical system with time-driven switching. We treat $N$ as a fixed number in this section, but relax this assumption later and handle a time-varying number of humans via replanning.

The robot plans its control actions over a finite horizon $[t_0, t_T]$ while the humans make stochastic transitions $\{y^i_k\}_{(1 \leq i \leq N, 1 \leq k \leq T)}$. Stacking these transition variables, we obtain a random vector with distribution $\mathcal{D}$. This distribution can be dependent on other variables such as past history of interactions and/or future motion plans of the robot. For the sake of planning, we only require that samples drawn from $\mathcal{D}$ are available. In this work we choose to model $\mathcal{D}$ with Trajectron++~\cite{salzmannivanovic2020trajectron++}, due to its superior performance over other trajectory forecasting methods as well as its capability to produce robot-future-conditional predictions. The reader is referred to \cite{salzmannivanovic2020trajectron++} for further details on Trajectron++.

\subsection{Optimal Control Problem}
Consider a finite horizon optimal control cost of the form
\begin{align}
    \label{eq: cost functional}
    J = \int_{t_0}^{t_T} c(s(t), u(t)) dt + h(s(t_T)),
\end{align}
where $c(\cdot)$ is the instantaneous cost function and $h(\cdot)$ is the terminal cost function.
In this work we assume that $c(\cdot)$ is the well-known LQ tracking cost plus a collision cost term:
\begin{multline}
    \label{eq: instant cost}
    c(s(t), u(t)) = \frac{1}{2}(x(t) - r(t))^\mathrm{T}Q(x(t) - r(t)) + \\
                       \frac{1}{2}u(t)^\mathrm{T}Ru(t) + c_\text{col}(s(t)),
\end{multline}
where $Q = Q^\mathrm{T}\succeq 0$ and $R = R^\mathrm{T} \succ 0$ are weight matrices, and $c_\text{col}(\cdot) \geq 0$ is a collision penalty function that is continuously differentiable, bounded, and has bounded gradient with respect to $x$. The reference trajectory $r$ is assumed to be given, possibly from a high level global planner.
Similarly, the terminal cost $h(\cdot)$ is defined by
\begin{multline}
    \label{eq: terminal cost}
    h(s(t_T)) = \frac{\beta}{2}(x(t_T) - r(t_T))^\mathrm{T}Q(x(t_T) - r(t_T)) + \\
                     \beta c_\text{col}(s(t_T)),
\end{multline}
where $\beta \geq 0$ determines the relative weight between the terminal and instantaneous costs.

Having specified the cost functional $J$, we define the (risk-neutral) stochastic optimal control problem as follows.
\begin{problem}[Risk-Neutral Stochastic Optimal Control]
    \label{problem: risk_neutral_soc}
    \begin{equation*}
    \begin{aligned}
    & \underset{u}{\text{minimize}} & & \mathbb{E}_\mathcal{D}[J] \\
    & \text{subject to}
    & & \eqref{eq: robot_dynamics}, \eqref{eq: human_dynamics}~\forall i \in \{1,\dots,N\} ~\forall k \in \{0,\dots,T\} \\
    &&& x(t_0) = x_0, p^i(t_0) = p^i_0 ~\forall i \in \{1, \dots, N\} \\
    &&& u(t) \in \mathcal{U} ~\forall t \in [t_0, t_T)
\end{aligned}
\end{equation*}
\end{problem}

\vspace*{1mm}
\subsection{Entropic Risk Measure}
The formulation of Problem \ref{problem: risk_neutral_soc} ignores the safety-critical aspect of collision avoidance, as simply optimizing the expected value fails to take into account the shape of the distribution. A remedy is to introduce the following entropic risk measure with risk sensitivity parameter $\sigma > 0$:
\begin{align}
    \label{eq: entropic risk def}
    R_{\mathcal{D},\sigma} (J) \triangleq \frac{1}{\sigma}\log\left(\mathbb{E}_\mathcal{D}[e^{\sigma J}]\right).
\end{align}
As long as the robot trajectory $x$ is bounded for all admissible control, $J$ becomes a bounded random variable and $R_{\mathcal{D},\sigma}(J)$ is finite. It is known that this transformation approximately decouples the mean and variance:
\begin{align}
    \label{eq: entropic risk approx}
    R_{\mathcal{D},\sigma} (J) \approx \mathbb{E}_\mathcal{D}[J] + \frac{\sigma}{2} \text{Var}_{\mathcal{D}}(J)
\end{align}
for small $\sigma \text{Var}_{\mathcal{D}}(J)$ \cite{whittle2002risk}. The meaning of $\sigma$ is now clear; it is a parameter that determines how much we care about the variability of the cost in addition to the mean. Larger $\sigma$ increases risk sensitivity, while the risk-neutral objective $\mathbb{E}_{\mathcal{D}}[J]$ is recovered as $\sigma \rightarrow 0^+$.

Replacing the expectation in Problem \ref{problem: risk_neutral_soc} with \eqref{eq: entropic risk def}, we obtain the following risk-sensitive optimal control problem:
\begin{problem}[Risk-Sensitive Stochastic Optimal Control]
    \label{problem: risk_sensitive_soc}
    \begin{equation*}
    \begin{aligned}
    & \underset{u}{\text{minimize}} & & R_{\mathcal{D},\sigma}(J) \\
    & \text{subject to} &&\text{the same constraints as in Problem \ref{problem: risk_neutral_soc}}
\end{aligned}
\end{equation*}
\end{problem}

\vspace*{1mm}
\section{RISK-SENSITIVE SEQUENTIAL ACTION CONTROL}
\label{sec: rssac}

\subsection{Review of Stochastic Sequential Action Control}
\label{sec: stochastic sac}
Even solving Problem \ref{problem: risk_neutral_soc} is intractable due to potential non-convexity in $c_\text{col}$ and complexity in $\mathcal{D}$. An efficient, approximate MPC solution can be obtained via Stochastic SAC \cite{nishimura2018sacbp, nishimura2019sacbp}. In this framework, we seek the optimal perturbation of a given nominal control such that the expected value of the mode insertion gradient \cite{wardi2012algorithm, ansari2016sequential}, which quantifies local effects of the perturbation on the cost functional $J$, is minimized. We assume that the nominal control is an open-loop control schedule $u$ (although it is straightforward to extend our method to deal with closed-loop nominal control policies \cite{nishimura2018sacbp, nishimura2019sacbp}). The perturbed control is defined as
\begin{align}
    \label{eq: perturbed_control}
    u^{\epsilon}(t) \triangleq \begin{cases}
        v & \text{if}\;\; t \in (\tau - \epsilon, \tau] \\
        u(t) & \text{otherwise},
    \end{cases}
\end{align}
where $v \in \mathcal{U}$, $\tau \in (t_0, t_T)$, $\epsilon \geq 0$ are the perturbation parameters. The perturbation $u^\epsilon$ yields deterministic, perturbed state trajectory $x^\epsilon$ and cost $J^\epsilon$ under a specific sample from $\mathcal{D}$. The mode insertion gradient is the sensitivity of the cost $J$ to the perturbation duration $\epsilon$, with $v$ and $\tau$ fixed:
\begin{align}
    \frac{\partial_+ J}{\partial \epsilon} \bigg|_{\epsilon = 0}  \triangleq \lim_{\epsilon\rightarrow 0^+} \frac{J^\epsilon - J}{\epsilon}.
\end{align}
The value of the mode insertion gradient is given by
\begin{multline}
    \label{eq: mig}
    \frac{\partial_+ J}{\partial \epsilon} \bigg|_{\epsilon = 0}  = \frac{1}{2}v^\mathrm{T}Rv  + \rho(\tau)^\mathrm{T} H(x(\tau)) (v - u(\tau)) \\- \frac{1}{2}u(\tau)^\mathrm{T}Ru(\tau),
\end{multline}
where $x$ is the robot state trajectory induced by nominal control $u$, and $\rho$ is the adjoint variable matching the dimensionality of the robot state. Specifically, it follows the ODE:
\begin{multline}
    \label{eq: adjoint_equation}
    \dot{\rho}(t) = -\frac{\partial}{\partial x} c(s(t), u(t)) \\-
    \left(\frac{\partial}{\partial x} f(x(t)) + \frac{\partial}{\partial x}H(x(t))u(t) \right)^\mathrm{T}\rho(t),
\end{multline}
with boundary condition $\rho(t_T) = \frac{\partial}{\partial x}h(s(t_T))$. As the joint state $s(t)$ is a random vector due to stochasticity in human motion, so is $\rho(t)$ and \eqref{eq: adjoint_equation} is valid under the specific sample from $\mathcal{D}$. Taking the expectation yields the expected mode insertion gradient $\mathbb{E}_{\mathcal{D}}\left[\partial_+ J/\partial\epsilon|_{\epsilon=0}\right]$, which is obtained by replacing $\rho(\tau)$ in \eqref{eq: mig} with $\mathbb{E}_{\mathcal{D}}[\rho(\tau)]$.

\subsection{Generalized Mode Insertion Gradient}
\label{sec: generalized mig}
Under a weak regularity condition on the dynamics function \eqref{eq: robot_dynamics} and for the cost $J$ defined by \eqref{eq: cost functional}, \eqref{eq: instant cost}, \eqref{eq: terminal cost}, one can show\footnote{A more general and detailed analysis is provided in  \cite{nishimura2019sacbp} where we provide sufficient conditions for \eqref{eq: generalized mig} to hold.} that the robot trajectory $x$ is bounded for all admissible control, and that the perturbed cost $J^{\epsilon}$ is Lipschitz continuous in $\epsilon \geq 0$:
\begin{align}
    \label{eq: cost lipschitz continuity}
    |J^{\epsilon} - J| \leq \epsilon K,
\end{align}
under the same random sample from $\mathcal{D}$ for $J$ and $J^\epsilon$. The value of the constant $K$ is not dependent on the samples. Therefore, the dominated convergence theorem allows us to interchange the derivative and the expectation:
\begin{align}
    \label{eq: generalized mig}
    \mathbb{E}_\mathcal{D}\left[\frac{\partial_+ J}{\partial\epsilon}\bigg|_{\epsilon=0}\right] = 
    \frac{\partial_+ \mathbb{E}_{\mathcal{D}}[J]}{\partial\epsilon}\bigg|_{\epsilon=0}.
\end{align}
The right-hand side of \eqref{eq: generalized mig} is the mode insertion gradient generalized to the case of risk-neutral stochastic optimal control (i.e. Problem \ref{problem: risk_neutral_soc}). If the optimal value with respect to $v$ is negative for some $\tau$, then there exists a sufficiently small $\epsilon$ for which the perturbation defined by $(v, \tau, \epsilon)$ will reduce the expected cost. If instead it is zero, we can think of the nominal control as satisfying a local optimality condition.

\subsection{Extension to Entropic Risk Measure}
Sections \ref{sec: stochastic sac} and \ref{sec: generalized mig} have provided a summary of prior work \cite{nishimura2018sacbp, nishimura2019sacbp}. Now we are set to derive the generalized mode insertion gradient for the entropic risk measure $R_{\mathcal{D},\sigma} (J)$, which is a novel contribution of this paper.
\begin{lemma}
    \label{lemma 1}
    Suppose that the total cost $J$ satisfies \eqref{eq: cost lipschitz continuity}. Then, for $\sigma > 0$ the following relation holds:
    \begin{align}
        \label{eq: exponential mig}
        \mathbb{E}_\mathcal{D}\left[\frac{\partial_+ e^{\sigma J}}{\partial\epsilon}\bigg|_{\epsilon=0}\right] = 
    \frac{\partial_+ \mathbb{E}_{\mathcal{D}}[e^{\sigma J}]}{\partial\epsilon}\bigg|_{\epsilon=0}.
    \end{align}
\end{lemma}
\begin{proof}
    Let $J^\epsilon$ and $J$ be the perturbed and the nominal cost, respectively. We have
    \begin{align}
        |e^{\sigma J^\epsilon} - e^{\sigma J}| = \frac{|e^{\sigma(J^\epsilon - J)} - 1|}{e^{\sigma J}}
        \leq e^{\sigma |J^\epsilon - J|} - 1,
    \end{align}
    where we used $|e^x - 1| \leq e^{|x|} - 1$ and $e^{\sigma J} \geq 1$. Substituting \eqref{eq: cost lipschitz continuity} and dividing by $\epsilon > 0$, we obtain
    \begin{align}
    \frac{|e^{\sigma J^\epsilon} - e^{\sigma J}|}{\epsilon} \leq \frac{e^{\sigma K \epsilon} - 1}{\epsilon}
    \end{align}
    Let $g(\epsilon)$ denote the right-hand side of this inequality, which is a strictly monotone increasing function for $\epsilon > 0$. Now let $|f_\epsilon(J)|$ denote the left-hand side as a function of $J$ parameterized by $\epsilon$. Take some positive $\epsilon_0$ and a sequence $\{\epsilon_n\}_{n \geq 0}$ converging to $0^+$  as $n \rightarrow \infty$. It follows that
    $\forall n \geq 0,~|f_{\epsilon_n} (J)| \leq g(\epsilon_0)$. The dominated convergence theorem applies as $g(\epsilon_0)$ is a finite, deterministic constant:
    \begin{align}
       \mathbb{E}_{\mathcal{D}}[ \lim_{n \rightarrow \infty} f_{\epsilon_n}(J)] = \lim_{n \rightarrow \infty} \mathbb{E}_{\mathcal{D}}[f_{\epsilon_n}(J)],
    \end{align}
    which is equivalent to \eqref{eq: exponential mig}.
\end{proof}
\begin{theorem}[Mode Insertion Gradient of Entropic Risk]
    \label{theorem 1}
    Suppose that the regularity condition mentioned in Section \ref{sec: generalized mig} is met so the robot trajectory $x$ is bounded under admissible control and the cost $J$ satisfies \eqref{eq: cost lipschitz continuity}. Then, for fixed $v$ and $\tau$ the mode insertion gradient of the entropic risk measure $\frac{\partial_+}{\partial\epsilon}R_{D, \sigma}(J)|_{\epsilon = 0}$ exists and is given by
    \begin{multline}
        \label{eq: entropic mig}
        \frac{\partial_+}{\partial\epsilon}R_{D, \sigma}(J)\bigg|_{\epsilon=0} = \\ \frac{1}{2}v^\mathrm{T}Rv + \frac{\mathbb{E}_{\mathcal{D}}[e^{\sigma J} \rho(\tau)]^\mathrm{T}}{\mathbb{E}_{\mathcal{D}}[e^{\sigma J}]} H(x(\tau)) (v - u(\tau)) \\- \frac{1}{2}u(\tau)^\mathrm{T}Ru(\tau),
    \end{multline}
    where $J$ inside the expectations is the cost value under the nominal control $u$.
\end{theorem}
\begin{proof}
    As the robot trajectory is bounded, $J$ is a bounded random variable and the value of $R_{\mathcal{D}, \sigma}(J)$ is finite. The chain rule gives
    \begin{align}
        \frac{\partial_+}{\partial \epsilon} R_{\mathcal{D},\sigma}(J)\bigg|_{\epsilon=0} &= \frac{1}{\sigma \mathbb{E}_{\mathcal{D}}[e^{\sigma J}]}\frac{\partial_+ \mathbb{E}_{\mathcal{D}}[e^{\sigma J}]}{\partial\epsilon}\bigg|_{\epsilon=0} \\
        &= \frac{1}{\mathbb{E}_{\mathcal{D}}[e^{\sigma J}]}\mathbb{E}_\mathcal{D}\left[e^{\sigma J} \frac{\partial_+ J}{\partial \epsilon} \bigg|_{\epsilon = 0} \right],
    \end{align}
    where we also used Lemma \ref{lemma 1}. Substituting \eqref{eq: mig} and simplifying the terms complete the proof.
\end{proof}

The mode insertion gradient of entropic risk \eqref{eq: entropic mig} is a generalization of the stochastic mode insertion gradient \eqref{eq: generalized mig}. Indeed, for the risk neutral case (i.e. $\sigma = 0$) the two equations match. This enables us to extend the Stochastic SAC algorithm to risk-sensitive optimal control problems without changing the structure of the algorithm.

\subsection{RSSAC Algorithm}
The core of RSSAC is the following optimization problem, which substitutes Problem \ref{problem: risk_neutral_soc} (if $\sigma = 0$) and \ref{problem: risk_sensitive_soc} (if $\sigma > 0$).
\begin{problem}[Mode Insertion Gradient Optimization]
    \label{problem: entropic mig optimization}
    \begin{equation*}
    \begin{aligned}
    & \underset{v, \tau}{\text{minimize}} & & \frac{\partial_+}{\partial \epsilon} R_{\mathcal{D},\sigma}(J)\bigg|_{\epsilon=0} \\
    & \text{subject to}
    & & \tau \in (t_0 + t_\text{calc}, t_T) \\
    &&& v(\tau) \in \mathcal{U} ~\forall \tau \in (t_0, t_T),
\end{aligned}
\end{equation*}
\end{problem}
where $t_\text{calc}$ is a computation time budget.
The objective can be evaluated by Monte Carlo sampling the joint dynamics \eqref{eq: robot_dynamics}, \eqref{eq: human_dynamics} under the nominal control and backward integration of the adjoint dynamics \eqref{eq: adjoint_equation}. Fixing $\tau$, Problem \ref{problem: entropic mig optimization} is a quadratic minimization over $v$ under a convex constraint. The optimal value $v^*(\tau)$ can be obtained analytically for simple constraints such as a box constraint or a norm inequality constraint with a scaled identity matrix $R$. Optimizing $\tau$ is achieved by solving $v^*(\tau)$ multiple times and searching for the minimum value. There is only a finite number of $\tau$ to consider since in practice we use numerical integration, such as the Euler scheme with some discrete step size, to integrate the robot dynamics. The optimal mode insertion gradient is always non-positive, and if it is negative then there exists some $\epsilon > 0$ such that the entropic risk measure is reduced as a result of the control perturbation. The value of $\epsilon$ can be either specified or searched. If instead the mode insertion gradient is zero, we set $\epsilon=0$ and choose not to perturb.

The pseudo-code of the RSSAC algorithm is presented in Algorithm \ref{algo: rssac}. 
Importantly, the Monte Carlo sampling part is naturally parallelizable. Note that the risk sensitivity has no effect unless we draw at least $M = 2$ samples. The replanning happens at every $\Delta t_\text{r}$ seconds, allowing for a variable number of humans to be considered over time. The complexity of the algorithm is $\mathcal{O}(NMT)$ where $N$ denotes the number of humans, $M$ the number of samples, and $T$ the planning horizon.

\begin{figure}
\begin{minipage}{\dimexpr\linewidth}
\begin{algorithm}[H]
	\caption{Control Computation with RSSAC}\label{algo: rssac}
	\begin{algorithmic}[1]
		\INPUT Initial joint state $s(t_0)$, reference trajectory $r$, nominal control schedule $u(t)$ for $t \in [t_0, t_T]$.
		\OUTPUT Optimally perturbed control schedule $u^{\epsilon}$
		\For{$j$ = 1:$M$}
		\State Forward-simulate the joint dynamics \eqref{eq: robot_dynamics}, \eqref{eq: human_dynamics} while sampling human transitions $\{y^i_k\}_{(1 \leq i \leq N, 1 \leq k \leq T)}^j$ from $\mathcal{D}$
		\State Compute sampled cost $J^j$
		\State Backward-simulate adjoint robot trajectory $\rho^j$ with sampled human transitions \eqref{eq: adjoint_equation}
		\EndFor
		\State Compute Monte Carlo estimate: $\mathbb{E}_{\mathcal{D}}[e^{\sigma J}\rho] \approx \frac{1}{M}\sum_{j=1}^M e^{\sigma J^{j}}\rho^{j}$ and $\mathbb{E}_{\mathcal{D}}[e^{\sigma J}] \approx \frac{1}{M}\sum_{j=1}^M e^{\sigma J^{j}}$
		\State Solve Problem \ref{problem: entropic mig optimization}
		\State Specify $\epsilon$ or search by re-simulating the dynamics
		\State $u^{\epsilon} \leftarrow PerturbControlSchedule(u,v^*,\tau^*,\epsilon)$ \eqref{eq: perturbed_control}
		\State \textbf{return} $u^{\epsilon}$
	\end{algorithmic}
\end{algorithm}
\end{minipage}
\vspace{-3mm}
\end{figure}

\subsection{Implementation Details}
\label{sec: impl details}
In our implementation, we use the double integrator model for the robot's dynamics: $x(t) = (x_\text{p}(t), x_\text{v}(t))$, $\dot{x}_\text{p}(t) = x_\text{v}(t)$, $\dot{x}_\text{v}(t) = u(t)$. These dynamics are integrated using the Euler scheme with step size $\Delta t_\text{c} = 0.02[s]$. The cost weight matrices are $Q = Diag(0.5, 0.5, 0, 0)$ and $R = 0.2I_{2\times2}$. The collision cost is the following sum of Gaussian functions centered at each human:
\begin{align}
    c_\text{col}(s(t)) = \sum_{i = 1}^N \alpha\exp\left(-\frac{\Vert x_\text{p}(t) - p^i(t)\Vert_2^2}{2\lambda}\right),
\end{align}
with peak parameter $\alpha = 100$ and bandwidth parameter $\lambda = 0.2$. The relative weight of the terminal cost is set to $\beta = 0.1$. The reference trajectory $r$ is a straight line connecting the initial robot position to the goal position at a constant target speed, and is replanned whenever $\Vert x(t) - r(t)\Vert_2 > 2[m]$. The control constraint is $\Vert u(t) \Vert_2 \leq u_{\max}$ with $u_{\max} = 5.0, 2.5 [m/s^2]$ in the simulation and the real-world experiment, respectively. Monte Carlo sampling is parallelized on a GPU with sample size $M = 30$. The planning horizon is $4.8[s]$, which corresponds to $T = 12$ steps of human motion prediction with measurement interval $\Delta t_\text{o} = 0.4[s]$. 

In contrast to prior work \cite{nishimura2018sacbp, nishimura2019sacbp}, we searched for $\epsilon$ from $\{0, 1e^{-3}, 2e^{-3}, 4e^{-3}, 8e^{-3}, 1.6e^{-2}, 2e^{-2}, 4e^{-2}, 8e^{-2}\}[s]$ by re-simulating the perturbed dynamics. The nominal control $u$ is a simple MPC-style search algorithm and takes the form
\begin{align}
    u(t) = \begin{cases} 
        u_\text{s}(t) & \text{if}\;\; t \in [t_0 + t_\text{calc}, t_0 + t_\text{calc} + \Delta t_\text{o}] \\
        u_\text{pr}(t) & \text{otherwise},
    \end{cases}
\end{align}
where $u_\text{pr}$ is the perturbed control from the previous iteration, and $u_\text{s}$ is either the same as $u_\text{pr}$ or chosen from a set of constant control inputs $\{(a \cos(\theta), a \sin(\theta))\}$ with $a \in \{0.4u_{\max}, 0.8u_{\max}\}$ and $\theta \in \{0, \pi/4, \pi/2, \dots, 2\pi\}$. The best $u_\text{s}$ is chosen based on the evaluation of $R_{\mathcal{D}, \sigma}(J)$ for each nominal control candidate using the Monte Carlo samples. This nominal search is similar to the iterative update scheme presented in \cite{tzorakoleftherakis2018iterative} in that the previously-computed perturbation is used in the next iteration, but we insert $u_\text{s}$ prior to running Algorithm \ref{algo: rssac}. It can also be considered as a simplified version of a tree search with motion primitives \cite{schmerling2018multimodal} with only 17 control choices and tree depth 1. Note that if robot-future-conditional prediction is used, the distribution $\mathcal{D}$ is different for each candidate nominal control, allowing the robot to consider the effect of various future robot trajectories on human behavior. We found this simple nominal search to be effective compared to constant nominal control \cite{ansari2016sequential, tzorakoleftherakis2018iterative} while retaining low computational cost.

We have implemented all the control code in Julia and achieved real-time performance with replanning interval $\Delta t_\text{r} = t_\text{calc} = 0.1 [s]$.


\section{RESULTS}
\label{sec: results}


\subsection{Simulation Results}

\begin{figure*}[t]
\vspace{5mm}
\begin{center}
	\begin{tabular}{c}
		\begin{minipage}[t]{0.3\textwidth}
			\centering
			\scalebox{1.0}[1.0]{\includegraphics[clip,width=1.0\columnwidth]{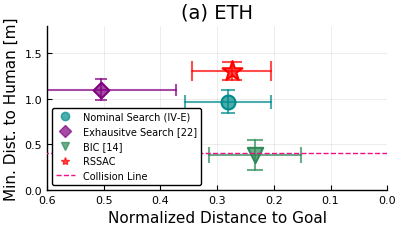}}
		\end{minipage}
		\begin{minipage}[t]{0.3\textwidth}
			\centering
			\scalebox{1.0}[1.0]{\includegraphics[clip,width=1.0\columnwidth]{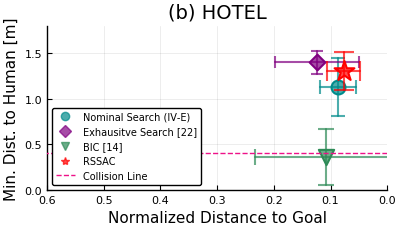}}
		\end{minipage}
		\begin{minipage}[t]{0.3\textwidth}
			\centering
			\scalebox{1.0}[1.0]{\includegraphics[clip,width=1.0\columnwidth]{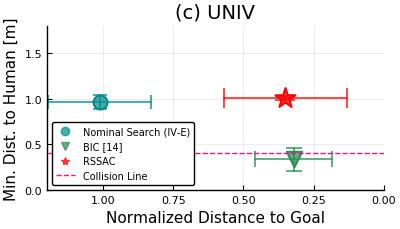}}
		\end{minipage}
	\end{tabular}
	\caption{Quantitative results from 100 runs show that risk-neutral (i.e. $\sigma = 0$) RSSAC further improves the performance of Nominal Search Only as the theory suggests, achieving both safety and efficiency. Note that the farther up and to the right, the better, as the x-axis is flipped. Exhaustive Search could not scale to the UNIV scene with more than 50 humans. BIC resulted in multiple collisions. Error bars show standard deviation.}
	\label{fig: data_benchmark_plot}
\end{center}
\end{figure*}
\begin{figure*}[t]
\begin{center}
	\begin{tabular}{c}
		\begin{minipage}[t]{0.3\textwidth}
			\centering
			\scalebox{1.0}[1.0]{\includegraphics[clip,width=1.0\columnwidth]{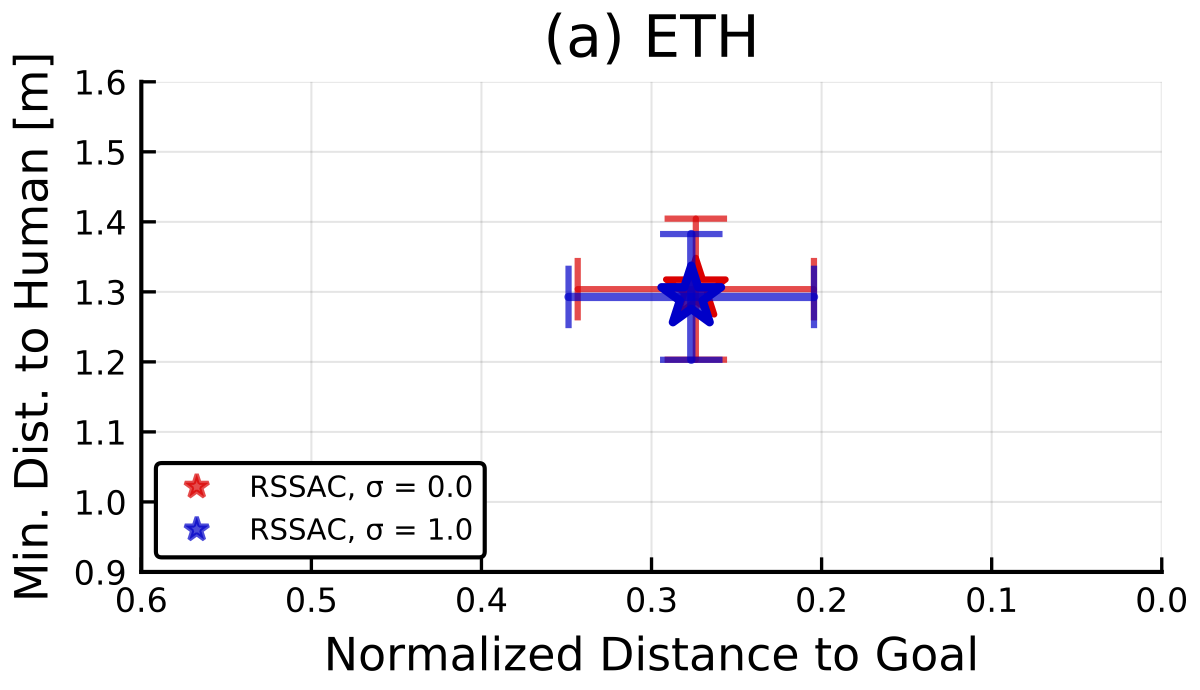}}
		\end{minipage}
		\begin{minipage}[t]{0.3\textwidth}
			\centering
			\scalebox{1.0}[1.0]{\includegraphics[clip,width=1.0\columnwidth]{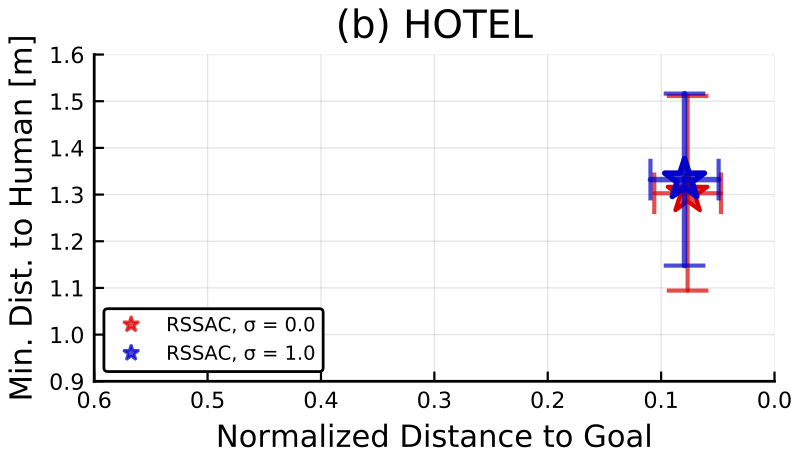}}
		\end{minipage}
		\begin{minipage}[t]{0.3\textwidth}
			\centering
			\scalebox{1.0}[1.0]{\includegraphics[clip,width=1.0\columnwidth]{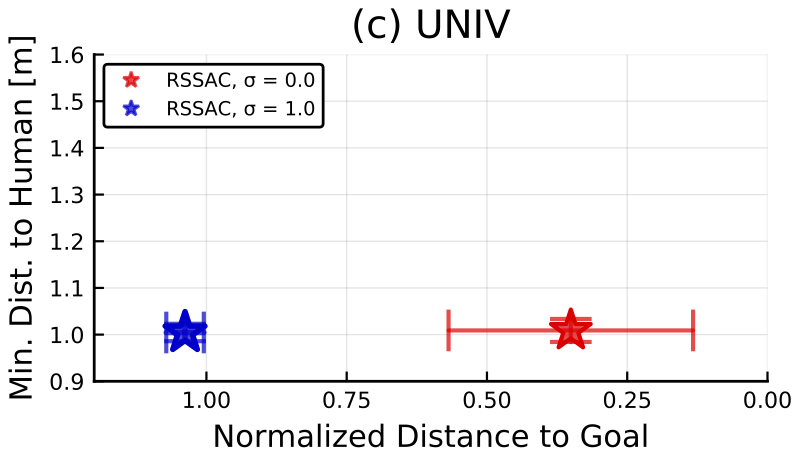}}
		\end{minipage}
	\end{tabular}
	\caption{Compared to the risk-neutral case in Fig. \ref{fig: data_benchmark_plot}, RSSAC with $\sigma = 1.0$ significantly reduces the standard deviation of the minimum robot-human distance by 11\%, 12\%, and 24\% in (a), (b), and (c), respectively. The risk sensitivity trades off the stochastic collision cost and the deterministic tracking cost, which results in increased standard deviation in the x-axis in (a) and (b), and overall distance increase in (c) where the scene was most densely-populated.}
	\label{fig: data_rs_plot}
\end{center}
\end{figure*}

We evaluated the performance of RSSAC in simulation and compare against three baseline collision avoidance algorithms: 1) LQ-Tracking  with Buffered Input Cell (BIC) \cite{wang2018safe}, 2) Nominal Search (Section \ref{sec: impl details}) Only, 3) Exhaustive Tree Search with Motion Primitives \cite{schmerling2018multimodal}. BIC is a reciprocal collision avoidance method that has similar computational complexity to the Velocity Obstacle (VO) approaches. Unlike VO, BIC does not require the velocity information of other agents to be available and is applicable to high-order linear dynamics. As our robot is a double-integrator and only obtains position measurements, BIC is a more suitable reciprocal collision avoidance approach than VO. BIC is minimally invasive in that an arbitrary control input is projected onto a convex polygon in the control space. Collision avoidance is guaranteed as long as all the agents respect their own BIC constraint. We use the LQ-Tracking cost (without $c_\text{col}$) to solve a deterministic optimal control problem and the BIC constraint adjusts the control input based on the current positions of humans. The robot is said to be in collision if it is within $40 [cm]$ from a human, but we set this threshold to $80 [cm]$ for planning with BIC only to give it an extra safety margin. The exhaustive search is similar to the nominal search in that it uses samples from $\mathcal{D}$, but the tree depth is $T = 4$ and at each depth the constant control is chosen from $\{(a \cos(\theta), a \sin(\theta))\}$ with $a \in \{0, 0.6u_{\max}\}$ and $\theta \in \{0, \pi/4, \pi/2, \dots, 2\pi\}$. This results in $9^4 = 6561$ sequences of control inputs to be considered. Despite parallelization on a GPU and a long replanning interval of $\Delta t_\text{r} = t_\text{calc} = 0.4[s]$, Exhaustive Search never achieved real-time performance. RSSAC, Nominal Search, and Exhaustive Search all used $M = 30$ prediction samples (per human) drawn from the Trajectron++ model \cite{salzmannivanovic2020trajectron++}.

For this evaluation, we used three distinct sequences from the publicly available ETH \cite{pellegrini2009you} and UCY \cite{lerner2007crowds} pedestrian motion datasets. They consist of real pedestrian trajectories with rich multi-human interaction scenarios. Each sequence that we used is a series of consecutive frames clipped from the ETH/ETH Test scene, the ETH/HOTEL Test scene, and the UCY/UNIV Test scene. For brevity, we refer to these as the ``ETH," ``HOTEL," and ``UNIV" sequences, respectively. In each scene, the robot starts from a collision free configuration and moves towards a specified goal while avoiding collisions. The ETH sequence is $10 [s]$ long and has 16 total pedestrians. The HOTEL sequence is also $10 [s]$ long and has 8 total pedestrians. The UNIV sequence is the most challenging with 95 total pedestrians over a horizon of $20 [s]$; there always exists 36 to 54 pedestrians simultaneously in each frame (see Fig. \ref{fig: demo}). Note that none of those pedestrians recognize or react to the robot as their motion is a replay from the data. This is the so-called ``invisible robot" setting \cite{chen2019crowd} where the robot has to predict Human-Human interaction only. For this reason, we did not condition the prediction of Trajectron++ on the nominal control candidates of the robot. 

Trajectron++ was trained following the same procedure as detailed in \cite{salzmannivanovic2020trajectron++}. The model was trained for 2000 steps with the Adam \cite{kingma2014adam} optimizer and a leave-one-out approach. That is, the model was trained on all datasets but the one used for evaluation. This training was done off-line prior to running RSSAC. At run time, RSSAC fetches prediction samples from the trained model every $\Delta t_\text{o} = 0.4[s]$ and uses them for the mode insertion gradient optimization. All the online computation was performed on a desktop computer with an Intel Xeon(R) CPU ES-2650 v3, 62.7 GiB memory, and a GeForce GTX 1080 graphics card.

Fig. \ref{fig: data_benchmark_plot} shows the statistical simulation results of RSSAC and the baseline methods. RSSAC, Nominal Search, Exhaustive Search were implemented with $\sigma = 0$. The x-axis shows the final distance between the robot and its goal normalized by the initial distance. The y-axis is the minimum distance between the robot and a human. These performance metrics measure the safety and efficiency of robot navigation, and are often in conflict. For each method we performed 100 runs with different random seeds. The goal of the robot was also randomized. As can be seen, RSSAC was both the safest and the most efficient among those methods that ran in real-time. BIC ended up in multiple collisions in all the three cases as the humans violated the reciprocity assumption. We did not test Exhaustive Search for the UNIV case due to its poor scalability; the computation took about $10 \times$ longer than the allocated time budget. We also note that RSSAC did further improve the performance of the Nominal Search, which itself was already achieving reasonably safe navigation.

\begin{figure}[t]
\vspace{5mm}
\begin{center}
	\begin{tabular}{c}
		\begin{minipage}[t]{0.45\columnwidth}
			\centering
 			\scalebox{1.0}[1.0]{\includegraphics[trim=40 0 0 0,clip,width=1.0\columnwidth]{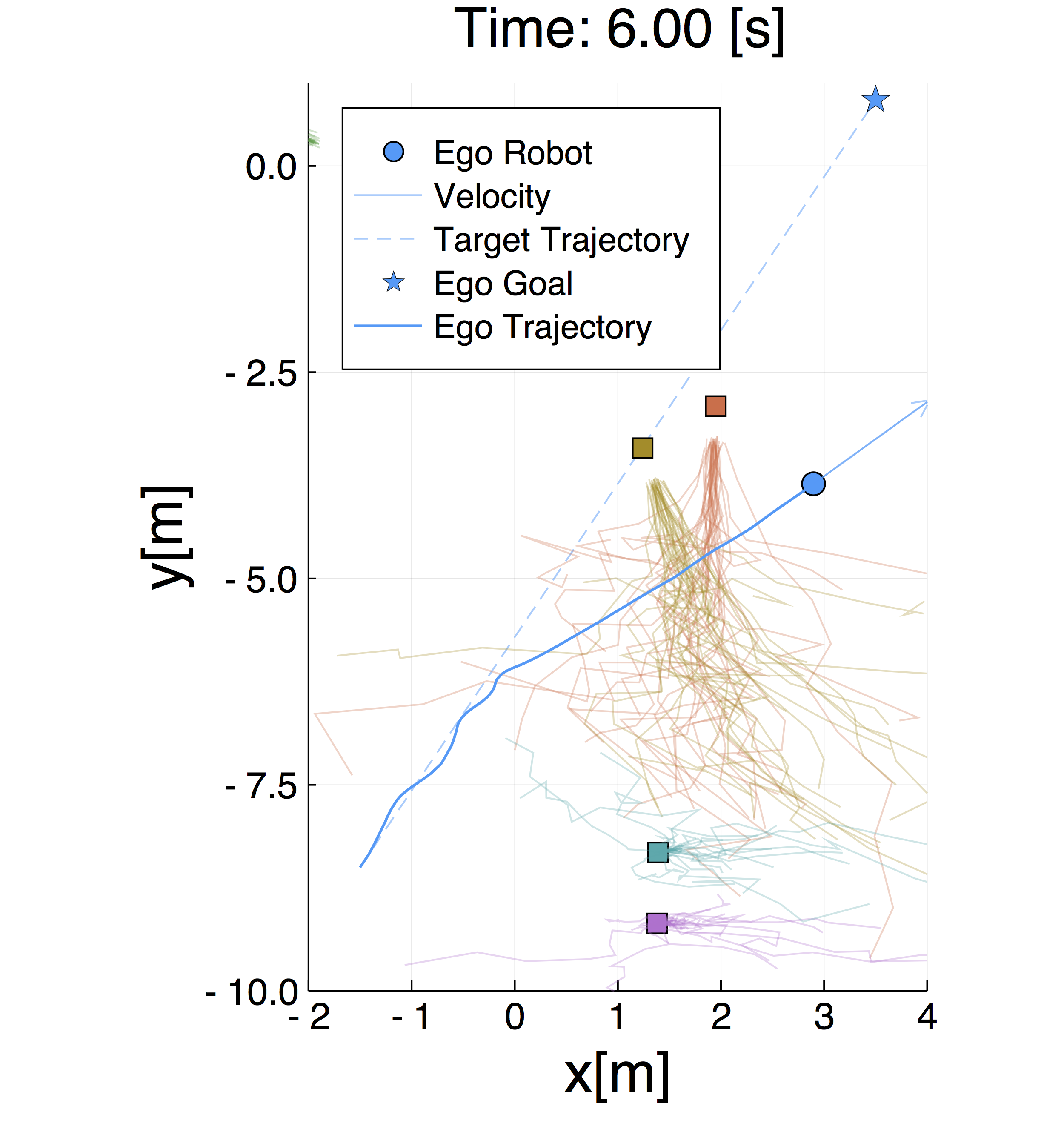}}
		\end{minipage}
		\begin{minipage}[t]{0.45\columnwidth}
			\centering
			\scalebox{1.0}[1.0]{\includegraphics[trim=40 0 0 0,clip,width=1.0\columnwidth]{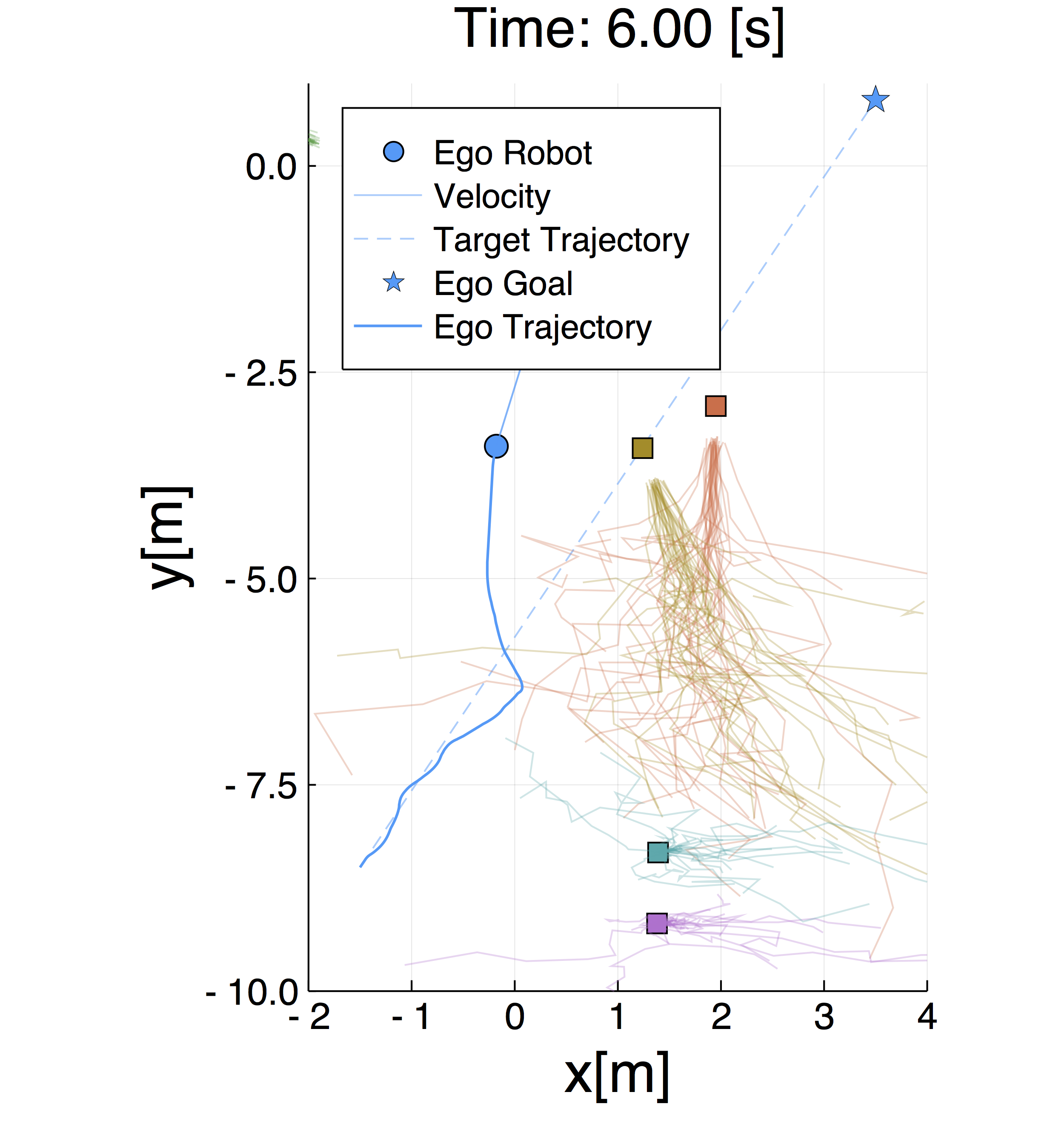}}
		\end{minipage}
	\end{tabular}
	\caption{Qualitative comparison of RSSAC with $\sigma = 0$ (left) and $\sigma = 1.0$ (right) in the HOTEL scene. These results differ in the minimum robot-human distance by only $3[cm]$ and the normalized goal distance by $0.01$, but the risk-sensitive robot (right) yields to potentially conflicting humans as opposed to the risk-neutral robot (left). Both simulations used the same random seed. Sampled predictions from Trajectron++ are also depicted.}
	\label{fig: hotel_risk_comparison_plot}
\end{center}
\vspace{-5mm}
\end{figure}

\subsection{Effects of Risk Sensitivity}

\begin{figure*}[t]
\vspace{5mm}
\begin{center}
	\begin{tabular}{c}
		\begin{minipage}[t]{0.25\textwidth}
			\centering
			\scalebox{1.0}[1.0]{\includegraphics[trim=30 0 0 35,clip,width=1.0\columnwidth]{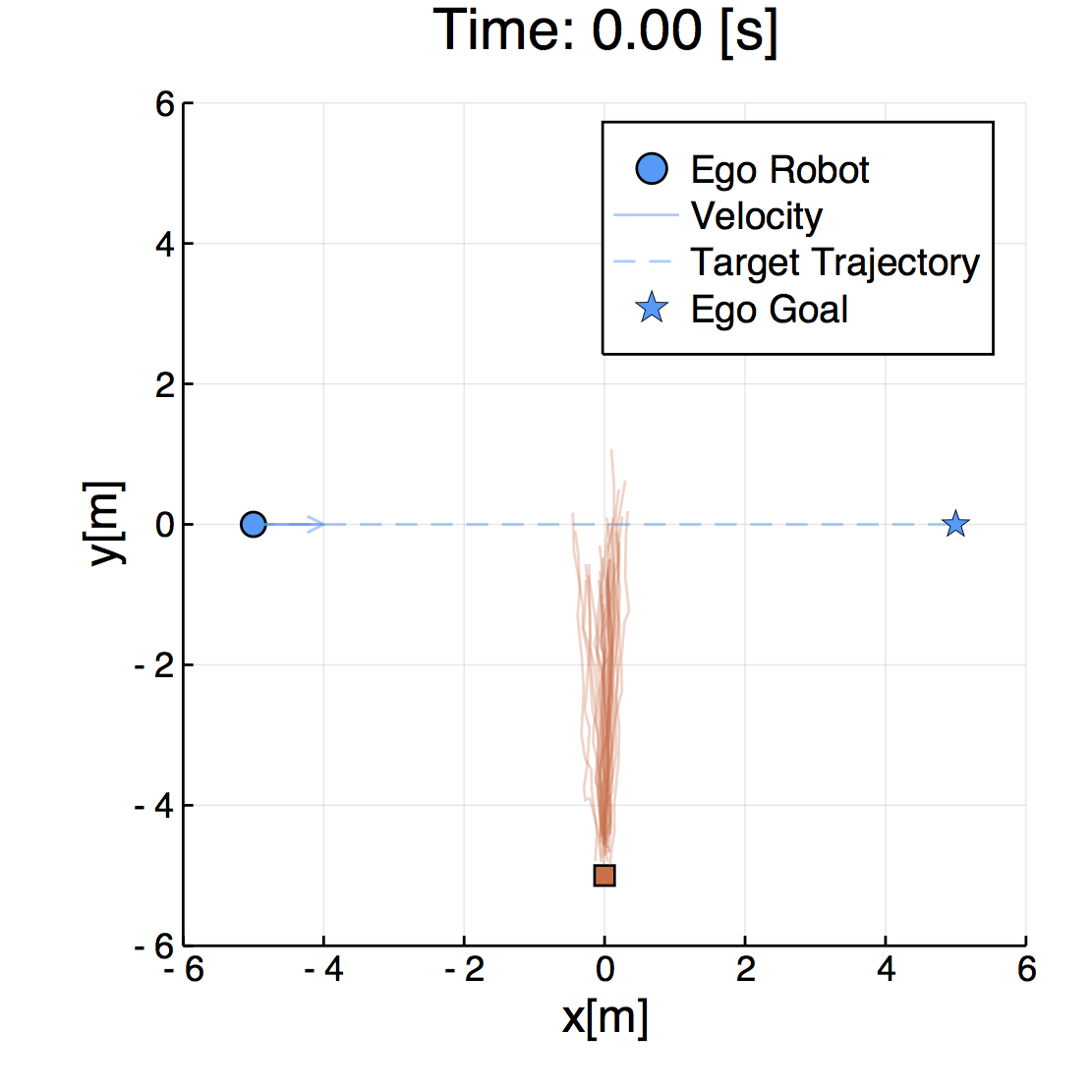}}
			\caption{A synthetic intersection scenario with a human. The prediction is drawn from a linear Gaussian model with a constant mean velocity.}
	        \label{fig: toy_intersection}
		\end{minipage}
		\hspace{0.5cm}
		\begin{minipage}[t]{0.32\textwidth}
			\centering
			\scalebox{1.0}[1.0]{\includegraphics[clip,width=1.0\columnwidth]{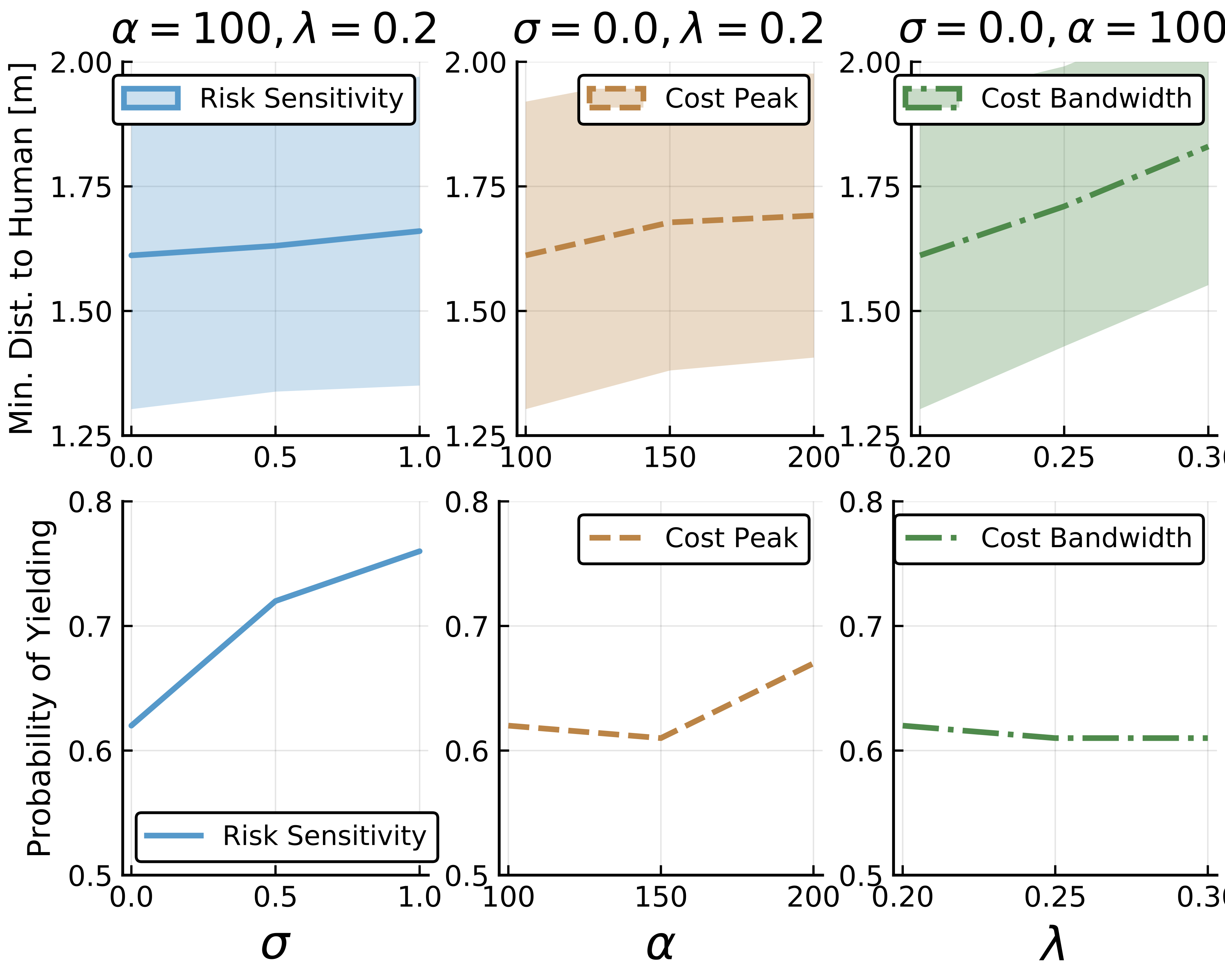}}
			\caption{Minimum robot-human distance (top) and empirical probability of yielding (bottom) for the synthetic intersection scenario. Changing the risk-sensitivity (left) consistently affected whether or not the robot yields, while the other two cost tuning parameters (middle and right) did not.}
	        \label{fig: gaussian_differences}
		\end{minipage}
		\hspace{0.5cm}
		\begin{minipage}[t]{0.33\textwidth}
			\centering
			\scalebox{1.0}[1.0]{\includegraphics[clip,width=1.0\columnwidth]{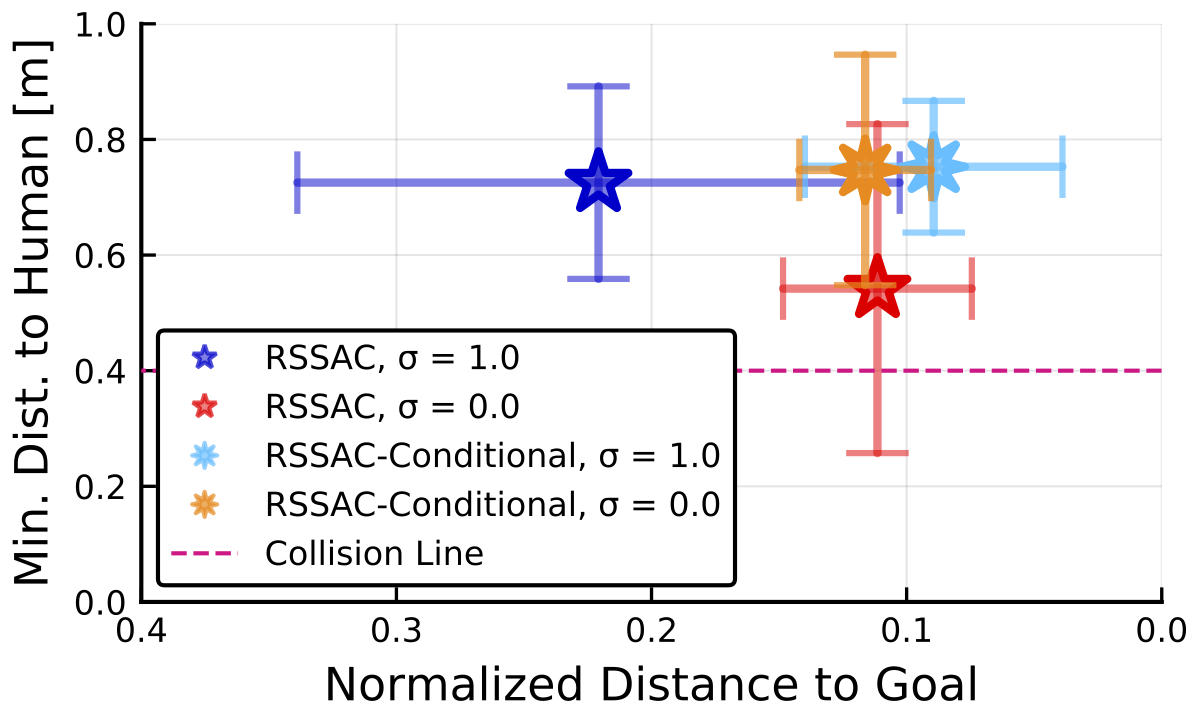}}
			\caption{Quantitative results of the real-world experiment with 5 human subjects. Using robot-future-conditional predictions with $\sigma = 1.0$ achieves the best average performance. Error bars show the standard deviation of 5 runs with a randomized robot goal and human start-goal assignment.}
	        \label{fig: real_experiment_plot}
		\end{minipage}
	\end{tabular}
\end{center}
\vspace{-5mm}
\end{figure*}

Next, we studied the effects of risk sensitivity in the context of safe robot navigation among humans. Fig. \ref{fig: data_rs_plot} compares RSSAC with $\sigma = 0$ (the same data as in Fig. \ref{fig: data_benchmark_plot}) and $\sigma = 1.0$. Our risk-sensitive optimization significantly reduced the standard deviation of the minimum robot-human distance by 11\%, 12\%, and 24\% for the ETH, HOTEL, and UNIV scenes, respectively. The risk sensitivity trades off the stochastic collision cost and the deterministic tracking cost, which appears in the ETH and HOTEL results where the standard deviation of the normalized goal distance slightly increased by 3.9\% and 2.3\%. On the other hand, the UNIV scene was so densely populated with humans that the risk-sensitive robot was unable to approach the goal. Although not necessarily captured by the quantitative metrics, overall the risk-sensitive robot also exhibited another intriguing behavioral difference as depicted in Fig. \ref{fig: hotel_risk_comparison_plot}; the risk-sensitive robot tends to yield to potentially conflicting humans.

To gain a better understanding of the effects of risk-sensitivity, we designed a simplified intersection scenario as illustrated in Fig. \ref{fig: toy_intersection}. The human in this toy example followed a linear Gaussian model with a constant mean velocity and was in a potential collision course with the robot. We ran 100 simulations for each $\sigma$ in $\{0.0, 0.5, 1.0\}$. The goal position was not randomized in this study. We also ran two additional sets of 100 simulations wherein we kept $\sigma = 0.0$ but varied the collision cost peak parameter $\alpha$ and the bandwidth $\lambda$, respectively. This was to elucidate the difference between risk sensitivity tuning and cost function design. The results are summarized in Fig. \ref{fig: gaussian_differences} and confirm that the more risk-sensitive the robot becomes, the more likely it is to yield to the human. This consistent trend was not present when changing $\alpha$ or $\lambda$ only. On the other hand, there was a positive correlation between the minimum robot-human distance and all the three parameters. This observation suggests that the risk sensitivity parameter $\sigma$ affects the global behavior of the robot that determines the ``type" of interaction, whereas cost function tuning affects the local behavior only (i.e. minimum robot-human distance). Thus, the risk sensitivity parameter can be considered as an additional degree of freedom in choosing desirable robot behavior.

\subsection{Real-World Experiment}
Our simulation study was accompanied by a real-world experiment with a holonomic robot \cite{wang2018ouijabots} and 5 human subjects in an indoor environment (see Fig. \ref{fig: demo}). Those subjects were assigned a specific start and goal positions, and were instructed to walk to each individual goal at normal speeds. Although the start-goal pairs remained the same, we changed their assignment to the subjects after each run so the type of interaction remained diverse. The robot started at a known position, but its exact goal was randomized and not known to the subjects. The positions of all the subjects as well as the robot were measured by a motion capture system. We used the same parameters as in the simulation study except $u_{\max}$. A major difference between the simulation and the real-world experiment is that the robot is ``visible" to humans in the experiment, which requires the robot to take into account resulting human-robot interaction in addition to human-human interaction. This was achieved by conditioning the prediction of Trajectron++ by the nominal control candidates for RSSAC, similar to \cite{schmerling2018multimodal}. As explained in Section \ref{sec: impl details}, this robot-future-conditional prediction lets the robot reason about the effect of different nominal control candidates on human behavior, prior to RSSAC perturbation on the best one. We compared the performance of the unconditional prediction (as in the simulation study) to the conditional prediction for both risk-neutral and risk-sensitive cases.

The results of the experiment are presented in Fig \ref{fig: real_experiment_plot}. For each setting we performed 5 runs. The robot with unconditional predictions was either too conservative $(\sigma = 1.0)$ or unsafe $(\sigma = 0.0)$. Using conditional predictions improved overall performance; we did not observe a single collision with conditional predictions. This supports our hypothesis that the robot-future-conditional prediction facilitates appropriate human-robot interaction. Of the four cases tested, using conditional predictions with $\sigma = 1.0$ achieved the best average performance, as well as the smallest variance in minimum robot-human distance.

\section{CONCLUSIONS}
\label{sec: conclusions}

This paper presents a novel online framework for safe crowd-robot interaction with dynamic collision avoidance. Our theoretical contribution is a derivation of the mode insertion gradient for the entropic risk measure. This theory leads to an efficient and high-performance implementation of RSSAC, the proposed  model predictive control algorithm for safety-critical systems. As a major practical contribution, we show that the probabilistic, robot-future-conditional predictions of Trajectron++ combined with the risk-sensitive optimization of RSSAC lead to safe and efficient robot navigation among many mutually-interacting humans. Furthermore, the risk sensitivity parameter is found to play a crucial role in determining the type of interaction behavior, such as yielding. In future work, we plan to apply the RSSAC-Trajectron++ framework to vehicle dynamics and scenes with heterogeneous agents. We are also interested in automatically adapting the risk sensitivity parameter so that any specifically-desired robot behavior emerges.

\bibliographystyle{IEEEtran}
\bibliography{IEEEabrv, references, asl_references, previous_asl_papers}

\end{document}